\documentclass{article}

\usepackage{amsmath}
\usepackage{amssymb}
\usepackage{adjustbox}
\usepackage{multirow}
\usepackage{wrapfig}

\usepackage{xcolor} 

\usepackage[preprint]{corl_2022} 

\title{NeuralGrasps: Learning Implicit Representations for Grasps of Multiple Robotic Hands}

\author{ Ninad Khargonkar$^1$ \hspace{3px} Neil Song$^2$ \hspace{3px} Zesheng Xu$^1$ \hspace{3px} Balakrishnan Prabhakaran$^1$ \hspace{3px} Yu Xiang$^{1}$\\
$^1$The University of Texas at Dallas \hspace{6px} $^2$St. Mark’s School of Texas\\
\tt\small \{ninadarun.khargonkar,zesheng.xu,bprabhakaran,yu.xiang\}@utdallas.edu \\ \tt\small 23songn@smtexas.org 
}

\begin{document}
\maketitle



\begin{abstract} We introduce a neural implicit representation for grasps of objects from multiple robotic hands. Different grasps across multiple robotic hands are encoded into a shared latent space. Each latent vector is learned to decode to the 3D shape of an object and the 3D shape of a robotic hand in a grasping pose in terms of the signed distance functions of the two 3D shapes. In addition, the distance metric in the latent space is learned to preserve the similarity between grasps across different robotic hands, where the similarity of grasps is defined according to contact regions of the robotic hands. This property enables our method to transfer grasps between different grippers including a human hand, and grasp transfer has the potential to share grasping skills between robots and enable robots to learn grasping skills from humans. Furthermore, the encoded signed distance functions of objects and grasps in our implicit representation can be used for 6D object pose estimation with grasping contact optimization from partial point clouds, which enables robotic grasping in the real world\footnote{Dataset and code are available at \url{https://irvlutd.github.io/NeuralGrasps}}.
\end{abstract}



\section{Introduction}
Robot manipulation is a fundamental research problem in robotics. If we want to have robots that can perform tasks to assist humans autonomously, we need to enable robots to grasp and manipulate objects and use tools to perform tasks. Robots have different grippers ranging from two-finger parallel grippers to five-finger anthropomorphic robotic hands. Usually, manipulation research typically focuses on one type of robot gripper. For instance, two-finger grippers are widely studied due to their simplicity in planning and control. Most commercial robots designed for research adopt two-finger grippers such as the Franka Emika Panda arm, the Fetch mobile manipulator, and the Baxter robot. As a result, manipulation skills learned from these robots are limited to two-finger grippers. It is unclear how to transfer or share these manipulation skills to robots with different grippers, especially for skills learned from imitating humans or reinforcement learning~\cite{mandlekar2018roboturk,wang2022goal}. Similarly, skills learned for multi-finger grippers are not transferable to two-finger grippers~\cite{andrychowicz2020learning}.

\begin{figure*}
	\centering
	\includegraphics[height=0.35\textwidth,width=\textwidth]{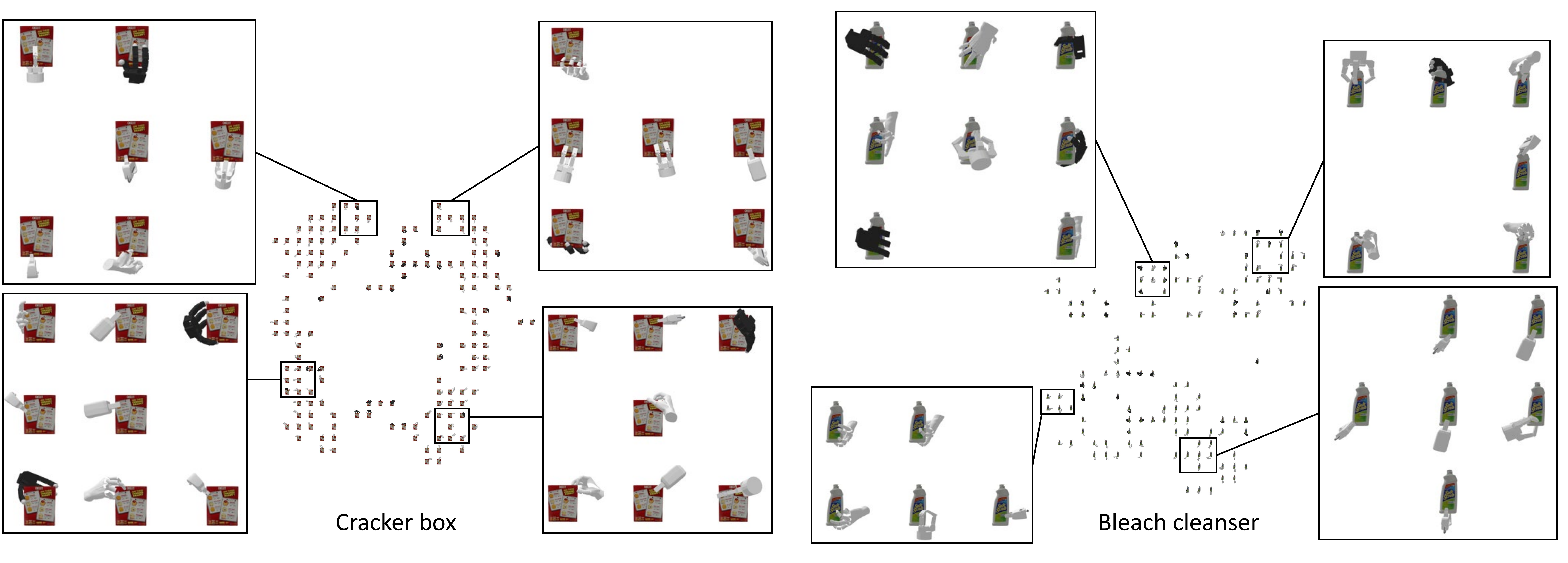}
	\caption{Barnes-Hut t-SNE visualization~\cite{van2014accelerating} using our learned embeddings of grasps from five different robotic hands.}
	\label{fig:t-sne}
	\vspace{4mm}
\end{figure*}

In this work, we study how to transfer grasps between different robot grippers. First, grasp transfer enables robots with different grippers to share grasping skills. Second, we consider the human hand to be a special robot gripper, so we can transfer human grasps to robot grippers. This will enable robots to learn grasping skills from human demonstrations. Learning from human demonstrations is very valuable in semantic grasping~\cite{dang2012semantic} or task-orientated grasping~\cite{kokic2020learning} where robots need to grasp objects according to the tasks. To achieve this goal, we introduce a novel implicit representation of multiple robotic hands learned using deep neural networks. Our representation learning is motivated by DeepSDF~\cite{park2019deepsdf} and Grasping Fields~\cite{karunratanakul2020grasping} which learn continuous Signed Distance Functions (SDFs) for 3D shapes of objects and human hands. We learn continuous SDFs for objects and multiple robotic hands. Our novelty lies in learning a latent space of grasps from multiple robotic hands, where a latent vector in this space encodes a grasp from a specific robotic hand. Importantly, the distance metric in the latent space encodes the similarity between grasps across different grippers. Therefore, given a grasp from one gripper, we can retrieve the closest grasp from another gripper using the distance metric in the latent space. This property enables grasp transfer between different grippers including human hands. Fig.~\ref{fig:t-sne} shows the t-SNE visualization of grasps from five different robotic hands using our learned latent vectors for two objects. We demonstrate grasp transfer from human hands to a Fetch mobile manipulator in our experiments.

Specifically, we use a deep neural network trained to encode an object and a number of grasps from multiple grippers. We use an object-centric view where all the grasps are defined with respect to the object coordinate frame. Given a latent vector $\mathbf{z}$ and a 3D point $\mathbf{x} \in \mathcal{R}^3$ as input, the network predicts the SDF values of the 3D point to the object and to the grasp of a gripper corresponding to $\mathbf{z}$. During training, in addition to the loss function on predicted SDF values, we utilize a triplet loss function~\cite{schroff2015facenet} to learn the distance metric in the latent space. In order to measure similarity between grasps from different grippers, we consider the contact maps of the grippers on the object, and use the similarity between contact maps to measure the grasp similarity. The triplet loss function extends the distance metric between contact maps to the latent space. In this way, the learned latent space preserves the similarity between grasps across different grippers.

Another advantage of our implicit representation is its use for real-world grasping. The learned SDFs of objects and grasps are fully differentiable functions. This property enables us to solve an optimization problem to estimate the 6D pose of object given a partial point cloud of the object as input.  During this optimization, we also consider the contact of a grasp and the point cloud of the object. With the estimated 6D object pose, a robot can use the encoded grasps to grasp the object. We demonstrated that optimizing contact between a grasp and the input point cloud can improve grasping success rate in the real world.

In summary, our contributions in this work are i) Presenting a novel implicit representation for grasps of multiple robotic hands; ii) Our representation enables grasp transfer between different robot grippers and human hands; iii) The representation can be used to solve object pose and optimize contact between grasps and observed point clouds in real-world grasping; iv) Novel grasping dataset with multiple robotic hands is introduced for multi-gripper transfer and to encourage future research.

\section{Related Work}
\label{sec:related-work}
\textbf{Neural Implicit Representations.} Neural implicit representations for scenes and objects have attracted widespread attention recently. Occupancy networks~\cite{mescheder2019occupancy} learn continuous occupancy functions of 3D shapes using neural networks. DeepSDF~\cite{park2019deepsdf} trains neural networks to encode the Signed Distance Functions (SDFs) of 3D shapes, while Neural Radiance Fields (NeRFs)~\cite{mildenhall2020nerf} encode both the geometry and color information of scenes into neural networks. The advantages of such implicit representations are: i) they are compact. For instance, a single DeepSDF network can encode a couple thousands of 3D shapes from the same category. NeRFs can encode large scenes with details with a single network. ii) The implicit representations are differentiable. So they can be easily used in gradient-based optimization to solve downstream tasks. For example, DeepSDF can be used for 3D reconstruction from partial observations. Grasping Fields~\cite{karunratanakul2020grasping} can synthesize humans grasps and reconstruct hands and objects jointly. The recently proposed Neural Descriptor Fields~\cite{simeonov2021neural} learn feature descriptors of 3D shapes for robot manipulation. The key difference between our work and related work is the implicit representation learning of grasps from multiple robotic hands and using the representation for both grasp transfer across grippers and object pose estimation for grasping.

\textbf{Grasping with Multiple Robotic Hands.} Traditional grasp planning methods such as the Graspit! simulator~\cite{miller2004graspit} use analytic approaches to access the grasp quality. These grasp quality measurements usually employ task wrench space analysis~\cite{ferrari1992planning,borst2004grasp} or force closure analysis~\cite{nguyen1988constructing,berenson2008grasp}. Analytic grasp planning methods can deal with different objects and different robotic hands. However, the main limitation of these traditional approaches is that they require full state information about objects such as shape and pose. They cannot work with partial observations of objects, e.g., point clouds from RGB-D cameras. Recent data-driven approaches for grasp planning utilize large-scale datasets with planned grasps~\cite{goldfeder2009columbia,eppner2021acronym} and machine learning techniques to learn models that can work with partial observations~\cite{bohg2013data,mahler2017dex,chu2018real,mousavian20196,sundermeyer2021contact}. However, majority of these works only focus on one type of robotic hand, especially, the two-finger parallel gripper. An exception is the UniGrasp~\cite{shao2020unigrasp} that considers gripper attributes in learning to detect grasping contact points on objects with a neural network. It can handle multiple grippers in detecting contact points for grasping. Our work differs from UniGrasp in that our implicit representations of objects and grasps enable grasp transfer between grippers and object pose estimation with grasp contact optimization. Another work closely related to ours is the ContactGrasp~\cite{brahmbhatt2019contactgrasp} that uses explicit contact maps to transfer grasps between human hands and robotic grippers. In contrast, our approach learns contact maps implicitly and uses a common latent space of grasps from human hands and robotic grippers for grasp transfer.

\section{NeuralGrasps}
\label{sec:method}
Our goal is to learn representations over different robotic hands that can be used for downstream tasks such as grasping and grasp transfer. We leverage recent progress in neural implicit representations for representing 3D shapes. Specifically, DeepSDF~\cite{park2019deepsdf} models the geometry of an object $o$ by learning the signed distance function \(f_{\text{SDF}}(\mathbf{x} ; o) : \mathcal{R}^3 \to \mathcal{R}\) over the object. 
Consequently, the signed distance function implicitly represents the surface points via the zero level set. DeepSDF models the 3D shapes of a set of objects by introducing a latent code for each object: \(f_{\text{SDF}} (\mathbf{x}, \mathbf{z}_i)\), where $\mathbf{z}_i \in \mathcal{R}^d$ with $d$ the dimension of the latent space and $i=1,2,\ldots,N$ for $N$ shapes. 




\subsection{Grasp Encoding Network}
\begin{figure*}
	\centering
	\includegraphics[width=0.95\textwidth]{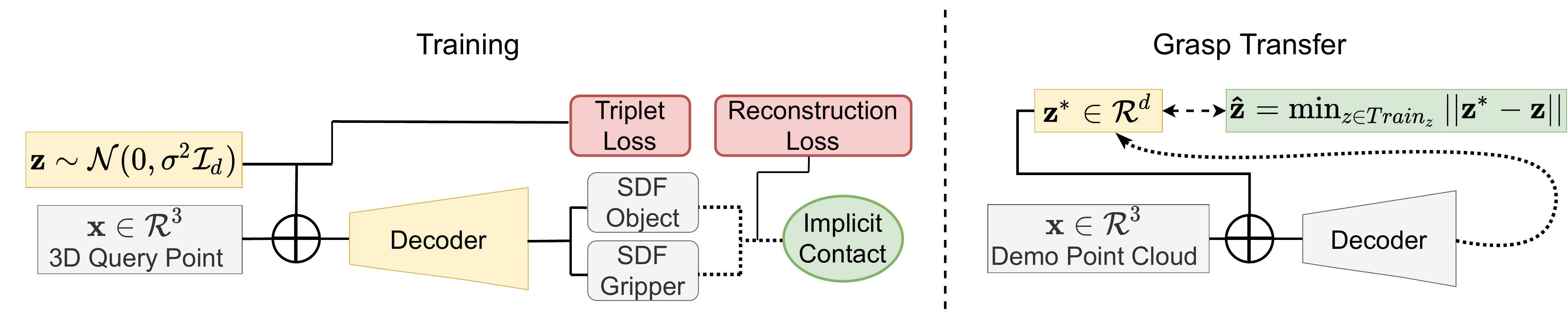}
	\caption{Left) Our network architecture with the training losses. Right) Illustration of the grasp transfer process during inference. Yellow modules indicate trainable parameters.}
	\label{fig:network-arch}
	\vspace{2mm}
\end{figure*}

Given a dataset of grasps from multiple grippers over an object \( o\), we represent each grasp between a gripper and the object via the signed distance functions of both the gripper and the object in a normalized, object-centric 3D space. Motivated by DeepSDF~\cite{park2019deepsdf} and Grasping Fields~\cite{karunratanakul2020grasping}, we jointly model the two signed distance functions using a \textit{grasp encoding} network. We utilize an auto-decoder based formulation as in DeepSDF where the network output is conditioned upon a latent vector \(\mathbf{z} \in \mathcal{R}^d\) corresponding to each grasping scene in the dataset. So the grasp encoding network models the following function: 
$f_{\text{SDF}} ( \mathbf{x}, \mathbf{z}; \theta_o) :  \mathcal{R}^3 \times \mathcal{R}^d \to \mathcal{R}^2$, where \(\theta_o\) denotes the network parameters corresponding to the model for object \(o\). The output of the network is two SDF values for the query point $\mathbf{x}$: one for the object $f_{\text{SDF}}^o ( \mathbf{x}; \mathbf{z}, \theta_o)$ and the other one for the gripper $f_{\text{SDF}}^g ( \mathbf{x}; \mathbf{z}, \theta_o)$. The network architecture is illustrated in Fig.~\ref{fig:network-arch}.
Note that for a given grasp encoded by its latent code \(\mathbf{z}\), the contact point set \(\mathcal{C}\) between the gripper and the object is implicitly represented by the zero level set: $\mathcal{C} = \{ \mathbf{x} \in \mathcal{R}^3 \;|\;  f_{\text{SDF}}(\mathbf{x}, \mathbf{z} ; \theta_o) = \mathbf{0} \}.$
Because these are 3D points both on the surface of the gripper and on the surface of the object. 


\subsection{Grasp Similarity}
\label{sec:grasp-similarity}


We aim to constrain the latent vector \(\mathbf{z}\) for a grasp to be close to the latent vectors of similar grasps. Since robot grippers have different kinematics, it is non-trivial to measure grasp similarity from different grippers. We consider an object-centric formulation here. We rely on the contact regions on the object to establish a similarity/distance measure over the grasp set. This follows the intuition that two \textit{similar} grasps probably interact with similar regions of the object. 

\begin{figure*}[!ht]
	\centering
	\includegraphics[height=0.22\textwidth, width=0.95\textwidth]{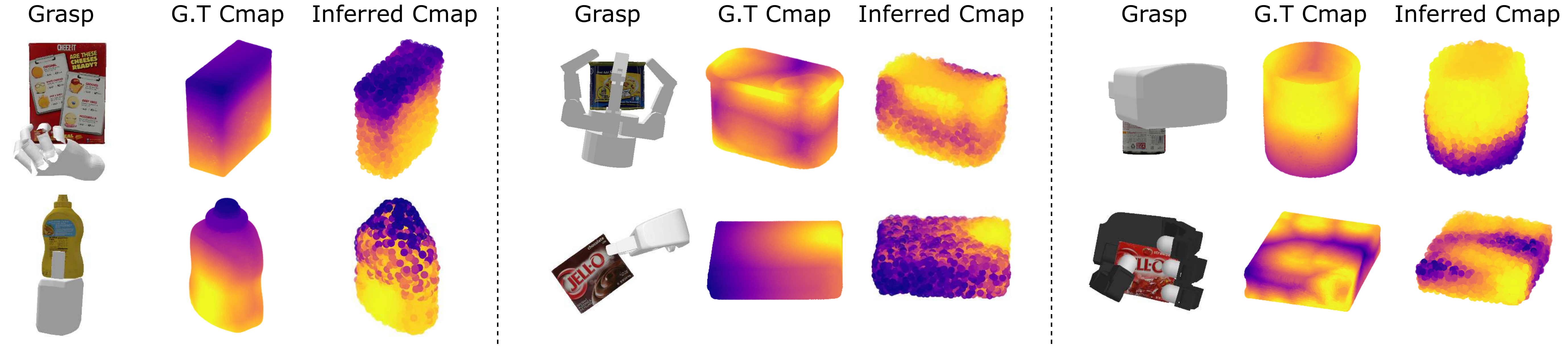}
	\caption{Illustration of several ground truth and inferred contact maps. Brighter regions correspond to the contact regions.}
	\label{fig:validation-contactmap}
\end{figure*}

\textbf{Contact Map Representation.} To compare grasps on the basis of their contact regions on the object, we compute a contact map \(\phi\) over the object points given a grasp. Given a grasping scene with two point clouds \(P_o, P_g \subset \mathcal{R}^{3}\) representing the object points and the gripper points, respectively. Let \(d(p_o, P_g)\) denote the distance between a point \(p_o \in P_o\) to the closest point in \(P_g\), i.e., \(d(p_o, P_g) = \min_{p_g \in P_g} \; d(p_o, p_g)\), where we use the standard \(L_2\) distance for \(d(p_o, p_g)\) in our experiments. To obtain the object-centric contact map \(\phi\), we simply take \(\phi \in \mathcal{R}^{|O|}\) with each dimension corresponding to an object point \(p_o \in P_o\). Then we define the contact map as \(\phi(p_o ; p_o\in P_o) = \exp(\frac{-d(p_o, P_g)}{\alpha}) \), where \(\alpha = 0.05m\) is a constant to penalize and score down object points with \(d(\cdot, P_g)\) value higher than \(\alpha\). Using this formulation, two grasps \(g_1, g_2\) are considered to be similar if \(\phi_{g_1} \sim \phi_{g_2}\). Note that the contact map is only assumed to be known at training time, and hence the unseen (test) grasps do not require a contact map for inference. A few representative visualization of contact maps across different grippers are shown in Fig.~\ref{fig:validation-contactmap}.

\subsection{Learning the Implicit Representations}
\label{sec:training-details}
For a given object $o$, we train the network parameters \(\theta_o\) on a dataset of grasps across multiple grippers. Each training sample \(T_i = \{\phi_i, X_i\},\; i=1,\dots N\), contains the contact map \(\phi_i\) for a grasp and a set of (query, SDF) pairs:
\(X_i = \{(\mathbf{x}_i^j, SDF(\mathbf{x}_i^j))_{j=1}^{M}  \;|\; \mathbf{x}_i^j \in \mathcal{R}^3 ; SDF(\mathbf{x}_i^j) \in \mathcal{R}^2 \}\), where $M$ indicates the number of query points and \(  SDF(\mathbf{x}_i^j) = (SDF_{o}(\mathbf{x}_i^j), SDF_{g}(\mathbf{x}_i^j))\) consists of the ground truth signed distances to the object and the gripper, respectively. Similar to the auto-decoder formulation in DeepSDF~\cite{park2019deepsdf}, the training proceeds by pairing each grasp (training sample) with a latent vector \(\mathbf{z}_i\) and minimizing the loss function which consists of the two terms described below. We assume a zero-mean Gaussian as the prior over the latent vectors.

\textbf{Reconstruction Loss.} The reconstruction loss $\mathcal{L}_{\text{SDF}}$ is the $L_1$ distance between the ground truth and the predicted SDF values for the gripper and the object over an input query point. A standard practice in learning SDFs is to clamp both the ground truth and predicted distances to be within a specific range $[ -\delta, \delta ]$ with a clamp function: $\text{clamp}(x, \delta) := \min(\delta, \max(-\delta, x))$, where $\delta$ is a parameter used to control the learning of SDF to be within a certain distance of the surface ($\delta=0.05m$ in our experiments). In this way, the learning focuses on the regions of the shape boundary. The reconstruction loss is defined as
\begin{equation}
\label{eqn:sdf-l1-loss}
	\mathcal{L}_{\text{SDF}}(\mathbf{x}, \mathbf{z}, \theta_o) = \vert \text{clamp}(  f_{\text{SDF}} (\mathbf{x}, \mathbf{z} ; \theta_o), \delta) - \text{clamp}( SDF(\mathbf{x}), \delta) \vert.
\end{equation}


\textbf{Triplet Loss in the Latent Space.} We constrain the latent vectors to represent a notion of similarity between the encoded grasps by utilizing the contact map \(\phi\). Our goal is to encourage \(\mathbf{z}_{g_1}, \mathbf{z}_{g_2}\) to be close to each other in the latent space if grasps \(g_1, g_2\) are similar to each other on the basis of their contact maps \(\phi_{g_1}, \phi_{g_2}\). We explicitly model such a constraint over the latent vectors during training since we use an encoder-less architecture. We make use of the triple loss~\cite{schroff2015facenet} with a variable margin for each triplet of latent vectors \((\mathbf{z}_a, \mathbf{z}_p, \mathbf{z}_n)\). The triplet \((a, p, n)\) represents the anchor, the positive and the negative training samples. We use the similarity between contact maps to define positive and negative examples w.r.t an anchor example. The triplet loss is defined as
\begin{equation}
\label{eqn:loss-triplet}
\mathcal{L}_{\text{triplet}}(\mathbf{z}_a, \mathbf{z}_p, \mathbf{z}_n) = \max\{ d(\mathbf{z}_a, \mathbf{z}_p) - d(\mathbf{z}_a, \mathbf{z}_n) + m(a,p,n) \;,\; 0 \},
\end{equation}
where $d(\cdot, \cdot)$ is the $L_2$ distance and $m(a,p,n)$ is the margin. In the original triplet loss function, the margin is a constant. In our case, we define the margin according to the similarity between grasp contact maps. Therefore, we can induce the grasp similarity into the latent space. The variable margin is defined as \(m(a,p,n) = ||\phi_a - \phi_n||_2 - ||\phi_a - \phi_p||_2\) using $L_2$ distance between contact maps. Without loss of generality, given an anchor and two other grasps, we can select the one with smaller contact map distance as the positive example and the other one as negative example. Therefore, the margin $m(a,p,n)$ is always greater than 0 for every sampled triplet. For each training batch, we randomly sample a fixed number of triplets and compute the mean of the triplet losses.

\textbf{Training Optimization.} Overall, we solve the following optimization problem to estimate the network parameters $\theta_o$ and latent codes $\{ \mathbf{z}_i \}_{i=1}^N$ for a set of $N$ grasps from multiple grippers:
\begin{equation} \label{eq:joint_hand_object}
    \theta_o^*, \{ \mathbf{z}_i^* \}_{i=1}^N = \arg\min_{\theta_o, \{ \mathbf{z}_i \}_{i=1}^N} \Big [ \sum_{i=1}^N \sum_{j=1}^M 	\mathcal{L}_{\text{SDF}}(\mathbf{x}_i^j, \mathbf{z}_i, \theta_o) + \sum_{(a, p, n)}\mathcal{L}_{\text{triplet}}(\mathbf{z}_a, \mathbf{z}_p, \mathbf{z}_n) + \frac{1}{\sigma^2} \sum_{i=1}^N  \| \mathbf{z}_i \|_2^2 \Big ],
\end{equation}
Here, each grasping scene contains \(M\) points and $\sigma$ is the standard deviation of a zero-mean multivariate-Gaussian prior distribution on the latent codes $\{ \mathbf{z}_i \}_{i=1}^N$.

\subsection{Solving Downstream Tasks with Our Implicit Representations}
\label{sec:inference-grasp-retrieval}
\textbf{Shape Reconstruction.} Given a latent vector \(\mathbf{z}\) corresponding to a grasp, we can perform inference over the learned network to reconstruct the shape of the object and the shape of the gripper. This is achieved by querying a set of 3D points and then obtaining the object surface points \( P_o = \{ \mathbf{x} \in \mathcal{R}^3 \;|\; f_{\text{SDF}}^{o}(\mathbf{x}, \mathbf{z} ; \theta_o) = 0 \}\) and the gripper surface \( P_g = \{\mathbf{x} \in \mathcal{R}^3 \;|\; f_{\text{SDF}}^{g}(\mathbf{x},  \mathbf{z} ; \theta_o) = 0 \}\). The contact points between the grasp and the object are intersection of the two sets of points. In cases when the latent vector $\mathbf{z}$ is unknown but we observe a set of points with their SDF values $(\mathbf{x}^j , SDF(\mathbf{x}^j))_{j=1}^M$, we can solve the following optimization problem to estimate the latent code:
\vspace{2mm}
\begin{equation} \label{eq:opt_latent}
    \mathbf{z}^* = \arg \min_{ \mathbf{z}} \Big [ \sum_{j=1}^M 	\mathcal{L}_{\text{SDF}}(\mathbf{x}^j, \mathbf{z}, \theta_o)  + \frac{1}{\sigma^2} \| \mathbf{z} \|_2^2 \Big ].
\end{equation}
In the real world, the query points $\mathbf{x}^j$ are usually from depth sensors. We can approximate the SDF samples using a similar scheme as shown in \cite{park2019deepsdf} by sampling points at a small distance away from surface points along the normal. 
On obtaining the latent code $\mathbf{z}^*$, we can reconstruct the shapes of the object and the gripper. 

\textbf{Grasp Transfer.} Due to the triplet loss imposed over the latent space, the inferred latent code $\mathbf{z}^*$ for an unseen grasp also follows the notion of grasp similarity. This metric learning constraint allows us to retrieve a most similar grasp in the dataset on the basis of the nearest neighbor in the latent code space as shown in Fig.~\ref{fig:network-arch}. The nearest neighbor grasp is not constrained to be from the input gripper, and hence such a framework makes it useful in a grasp transfer scenario where the demo and target grippers might differ. For example, the input grasp can be from a human as grasping demonstration. 

\textbf{6D Object Pose Estimation with Grasp Contact.} Since our implicit representation encodes the SDFs of an object and its grasps, we can utilize it to estimate the 6D pose of the object, i.e., 3D rotation $R$ and translation $\mathbf{t}$, given camera observations in the real world. Suppose we can segment a set of 3D points of the object in the camera frame $\{ \mathbf{x}_c^j \}_{j=1}^M$. If we transform these points into the object coordinate frame using $(R, \mathbf{t})$, they should have SDF values of the object close to zeros, since the transformed points are on the object surface. In addition, if the final goal is to execute a grasp $g$ to grasp the object, we can optimize the object pose such that the contact points of grasp $g$ are close to the observed 3D points. Let $ \{ \mathbf{x}_g^k  \}_{k=1}^L \subset   \{ \mathbf{x}_c^j \}_{j=1}^M$ be the contact points of grasp $g$ in the camera frame. We solve the following optimization problem to estimate the object pose:
\begin{equation} \label{eq:6d-pose}
    R^*, \mathbf{t}^* = \arg \min_{ R, \mathbf{t}} \Big [ \sum_{j=1}^M  | f_{\text{SDF}}^o(R \mathbf{x}_c^j + \mathbf{t}, \mathbf{z}; \theta_o) | + \sum_{k=1}^L | f_{\text{SDF}}^g(R \mathbf{x}_g^k + \mathbf{t}, \mathbf{z}; \theta_o) |	 \Big ],
\end{equation}
where $f_{\text{SDF}}^o$ and $f_{\text{SDF}}^g$ are the SDF predictions of the object and the gripper from the network and $\mathbf{z}$ is the latent code for grasp $g$. Starting from an initial object pose, we can apply gradient descent to optimize object pose estimation. More details on object pose estimation are provided in the supplementary materials.

\section{Multi-Hand Grasping Dataset}\label{sec:dataset}
\begin{figure*}
	\centering
	\includegraphics[width=0.99\textwidth]{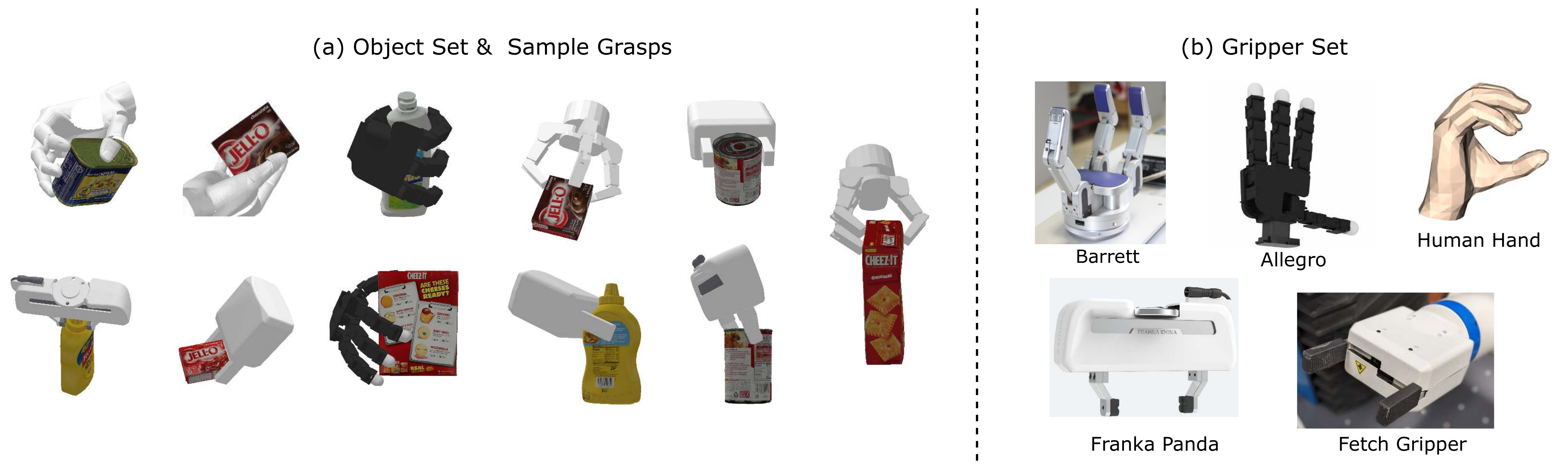}
	\caption{Objects and robotic hands in our multi-hand grasping dataset.}
	\label{fig:dataset-grippers-grasps}
\end{figure*}

Most of the current datasets for object manipulation exclusively focus either on parallel jaw grippers~\cite{eppner2021acronym} or human hand-object interaction \cite{hasson2019learning}. Such datasets lack the data needed for learning neural implicit representations of multiple grippers. In order to train our proposed model, we generated a synthetic dataset of common robot grippers grasping objects. Creating a synthetic dataset of grasps over 3D models of real world objects allows for high detail grasping scenes containing the dense point clouds, contact regions, and signed distance function values. 

\textbf{Grippers and Objects.} We selected five gripper models and seven objects from the YCB object set \cite{calli2015ycb-objectset}. The gripper set includes four robot grippers across a range of the number of fingers: 2-finger Fetch and Franka-Panda grippers, 3-finger Barrett gripper, 4-finger Allegro finger, and finally a human hand model. The Fetch gripper was also used in our real-world experiments. The motivation to include a diverse set of multi-fingered grippers was to investigate how well such an encoding scheme works across grippers with different kinematic configurations. The seven objects from the YCB set are cracker box, tomato soup can, mustard bottle, pudding box, gelatin box, potted meat can, and bleach cleanser. Fig.~\ref{fig:dataset-grippers-grasps} shows the objects and grippers grasps in our dataset.

\textbf{Grasp Generation.} To generate diverse multi-hand grasps over the YCB objects, we utilized the Graspit! simulator~\cite{miller2004graspit}. We initially generated a large number of grasps from GraspIt! and later sampled the initial set to generate a fixed amount of grasps for each gripper using the farthest point sampling algorithm which ensures diversity of the 3D position in the sampled grasp set. After the sampling stage, each grasping scene, i.e., an object with a gripper in a grasping pose, was loaded in the Pybullet \cite{coumans2016pybullet} environment, and dense point clouds were rendered for the grasping scene using multiple viewpoints.
Using the rendered point clouds, we generated the SDF values and the contact maps in the training samples as described in Section \ref{sec:training-details}. We followed DeepSDF~\cite{park2019deepsdf} to sample majority of SDF values on query points close to the gripper and object surface. Our pipeline for data generation can be easily extended with additional sets of objects and grippers.

\vspace{-2mm}
\section{Experiments}
\vspace{-2mm}
\label{sec:experiments}
\begin{table}[]
\caption{Chamfer distance ($\times$1e-3) for shape reconstruction and similarity scores for grasp retrieval}
\label{tab:exp-cdist-retrieval}
\centering
\resizebox{0.9\linewidth}{!}{\begin{tabular}{|c|cccccc||cccc|}
\hline
YCB Model &
  \multicolumn{6}{c||}{Shape Reconstruction} &
  \multicolumn{4}{c|}{Grasp Retrieval} \\ \hline
 &
  \multicolumn{1}{c|}{Object} &
  \multicolumn{1}{c|}{Fetch} &
  \multicolumn{1}{c|}{Barrett} &
  \multicolumn{1}{c|}{Human} &
  \multicolumn{1}{c|}{Allegro} &
  Panda &
  \multicolumn{1}{c|}{Near Z} &
  \multicolumn{1}{c|}{Near GT} &
  \multicolumn{1}{c|}{Far Z} &
  Far GT \\ \cline{2-11} 
\textit{Cracker Box} &
  \multicolumn{1}{c|}{1.33} &
  \multicolumn{1}{c|}{1.44} &
  \multicolumn{1}{c|}{5.52} &
  \multicolumn{1}{c|}{2.86} &
  \multicolumn{1}{c|}{3.03} &
  2.48 &
  \multicolumn{1}{c|}{0.81} &
  \multicolumn{1}{c|}{0.88} &
  \multicolumn{1}{c|}{0.19} &
  0.15 \\
\textit{Soup Can} &
  \multicolumn{1}{c|}{2.60} &
  \multicolumn{1}{c|}{4.95} &
  \multicolumn{1}{c|}{4.06} &
  \multicolumn{1}{c|}{3.21} &
  \multicolumn{1}{c|}{3.97} &
  4.20 &
  \multicolumn{1}{c|}{0.73} &
  \multicolumn{1}{c|}{0.84} &
  \multicolumn{1}{c|}{0.13} &
  0.09 \\
\textit{Mustard Bottle} &
  \multicolumn{1}{c|}{1.05} &
  \multicolumn{1}{c|}{2.38} &
  \multicolumn{1}{c|}{3.39} &
  \multicolumn{1}{c|}{2.05} &
  \multicolumn{1}{c|}{2.09} &
  4.14 &
  \multicolumn{1}{c|}{0.80} &
  \multicolumn{1}{c|}{0.87} &
  \multicolumn{1}{c|}{0.14} &
  0.10 \\
\textit{Pudding Box} &
  \multicolumn{1}{c|}{1.49} &
  \multicolumn{1}{c|}{5.27} &
  \multicolumn{1}{c|}{8.27} &
  \multicolumn{1}{c|}{5.45} &
  \multicolumn{1}{c|}{3.80} &
  5.40 &
  \multicolumn{1}{c|}{0.77} &
  \multicolumn{1}{c|}{0.84} &
  \multicolumn{1}{c|}{0.25} &
  0.16 \\
\textit{Gelatin Box} &
  \multicolumn{1}{c|}{1.20} &
  \multicolumn{1}{c|}{5.42} &
  \multicolumn{1}{c|}{7.04} &
  \multicolumn{1}{c|}{4.72} &
  \multicolumn{1}{c|}{3.74} &
  3.72 &
  \multicolumn{1}{c|}{0.75} &
  \multicolumn{1}{c|}{0.81} &
  \multicolumn{1}{c|}{0.22} &
  0.14 \\
\textit{Potted Meat Can} &
  \multicolumn{1}{c|}{2.13} &
  \multicolumn{1}{c|}{3.63} &
  \multicolumn{1}{c|}{3.78} &
  \multicolumn{1}{c|}{3.06} &
  \multicolumn{1}{c|}{4.45} &
  1.78 &
  \multicolumn{1}{c|}{0.76} &
  \multicolumn{1}{c|}{0.83} &
  \multicolumn{1}{c|}{0.21} &
  0.12 \\
\textit{Bleach Cleanser} &
  \multicolumn{1}{c|}{7.82} &
  \multicolumn{1}{c|}{2.45} &
  \multicolumn{1}{c|}{5.63} &
  \multicolumn{1}{c|}{7.22} &
  \multicolumn{1}{c|}{5.41} &
  14.08 &
  \multicolumn{1}{c|}{0.86} &
  \multicolumn{1}{c|}{0.91} &
  \multicolumn{1}{c|}{0.16} &
  0.13 \\ \hline
\end{tabular}}
\end{table}

\subsection{Representation Learning on the Multi-Hand Grasping Dataset}
To the best of our knowledge, there is no previous method that encodes grasps from multiple robotic grippers using neural implicit representation and enforces a notion of grasp similarity in a unified fashion. Hence, to validate our representation, we focus on the tasks of (1) shape reconstruction from a given latent code and, (2) grasp retrieval using the latent codes. 
\begin{figure*}
	\centering
	\includegraphics[height=0.2\textwidth, width=0.9\textwidth]{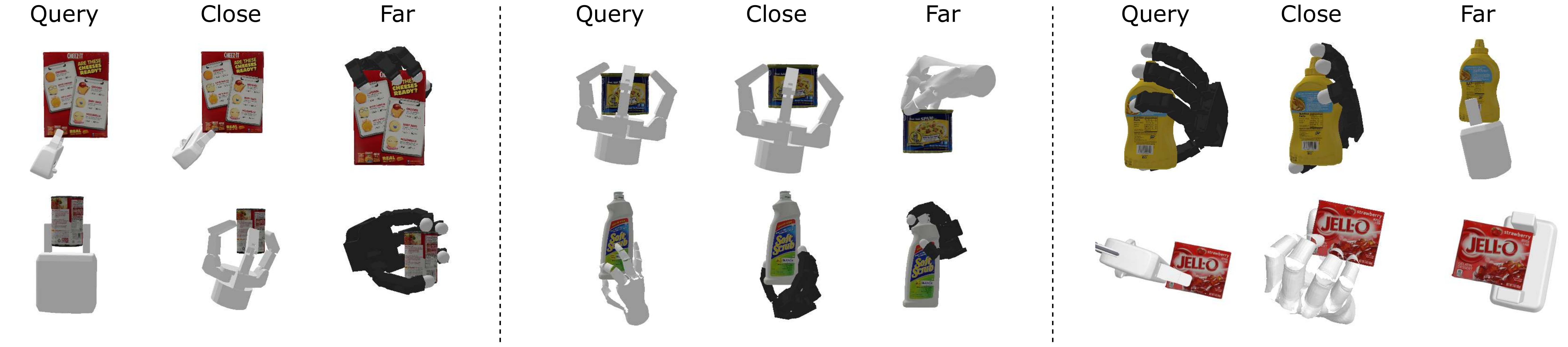}
	\caption{Grasp Retrieval: Query grasps from test set with close/far grasps from training set}
	\label{fig:grasp_retrieval}
	\vspace{6mm}
\end{figure*}

\textbf{Shape Reconstruction.} For this task, we consider a set of unseen test grasps with inferred latent codes from Eqn.~\eqref{eq:opt_latent}. For each test grasp, query points are randomly sampled and passed as inputs to the network with the specific latent code. Using the predicted SDF values on the set of query points, the object and gripper points are separated out (\(P_o , P_g\) in Section~\ref{sec:inference-grasp-retrieval}). We then compute the Chamfer distance against their ground truth point clouds and report the results in Table ~\ref{tab:exp-cdist-retrieval}. The results shown in the table indicate a good performance even though reconstruction is not the primary goal of the method and is only included as a validation step for the representation learning. Furthermore, the inferred contact maps for such reconstructions align closely with the ground truth as seen in Fig.~\ref{fig:validation-contactmap}.

\textbf{Grasp Retrieval.}
To test the effectiveness of the metric learning loss on the latent space, we perform an experiment on grasp retrieval for an input query grasp. We use the latent vector of an input query grasp and retrieve the closest and farthest grasps using (a) the distance between the latent vectors as a similarity measure, and (b) the distance between the contact maps which is considered as the ground truth (GT in Table~\ref{tab:exp-cdist-retrieval}). We report the similarity score \(s = 1-d \;; d \in  [0, 1]\) between the query and retrieved grasps for both (a) and (b) in
Table \ref{tab:exp-cdist-retrieval}. \(d\) is the normalized \(L_{1}\) distance between the contact maps of the pairs of grasps. As seen in Table \ref{tab:exp-cdist-retrieval}, the learned latent codes have similarity scores aligned with those for the contact maps based method, and this shows the effectiveness of the metric learning constraint. Fig.~\ref{fig:grasp_retrieval} shows some grasp retrieval examples using the latent codes.

\subsection{Real-World Experiments on Grasping and Grasp Transfer}
\textbf{Grasping with Object Pose Estimation.}
\label{sec:experiment-object-pose}
To validate our implicit representation in the real world, we first conducted a grasping experiment with the 7 YCB objects in our dataset on a tabletop. A Fetch mobile manipulator is employed for grasping. For each trial, we simply put one object on a table and we use the table height to segment the point cloud of the object from the Fetch head camera. Using the segmented point cloud, we estimate the object pose according to Eqn.~\eqref{eq:6d-pose}. For comparison, we also tested a baseline model  that only uses the SDF of the object without grasp contact optimization, i.e., no $f_{\text{SDF}}^g$ in Eqn.~\eqref{eq:6d-pose}. Since Eqn.~\eqref{eq:6d-pose} requires a grasp for optimization, we first use the baseline model to estimate the object pose, and then select the closest grasp from the grasp set to the current gripper location for optimization. Using the estimated pose, we attempt to grasp and lift the object after finding a suitable motion plan using MoveIt. We did 5 trials for each object. The results of this study are shown in Table ~\ref{tab:exp-refinment} where we can see the relative improvement in grasp success rate for the method using the contact-based pose estimation. This also shows the importance of the implicit contact point modeling in the framework since contact is a critical component of grasping. Please see the supplementary materials for the grasping videos.

\begin{wraptable}{r}{0.4\textwidth}
\caption{Grasp success over 5 trials: baseline vs. grasp contact optimization \vspace{0.5mm}}
\label{tab:exp-refinment}
\centering
\resizebox{\linewidth}{!}{\begin{tabular}{c|c|c}
\hline
YCB Model            & Baseline & With Contact \\ \hline
\textit{Cracker Box} & 5        & 5               \\
\textit{Soup Can}    & 2        & 4               \\
\textit{Mustard}     & 2        & 4               \\
\textit{Pudding Box} & 4        & 4               \\
\textit{Gelatin Box} & 3        & 5               \\
\textit{Potted Meat} & 3        & 4               \\
\textit{Bleach}      & 3        & 4               \\ \hline
\#Success & 22    & 30              \\ \hline
Success Rate         & 0.628    & 0.857           \\ \hline
\end{tabular}}
\end{wraptable}

\textbf{Human-to-Robot Grasp Transfer.}
\label{sec:experiment-human-demo}
We consider the task of transferring human grasps to our Fetch robot via the proposed implicit representation. We first conduct object pose estimation as described above. After the pose estimation, a person demonstrates a grasp on the object. Using the RGB-D images from the Fetch camera, we estimate the 3D hand joints using A2J~\cite{xiong2019a2j} and then utilize Pose2Mesh \cite{choi2020pose2mesh} to reconstruct the 3D hand mesh from the predicted 3D hand joints. We combine the hand points from the 3D hand mesh with the object point cloud and infer a latent code for the demonstrated grasp via Eqn.~\eqref{eq:opt_latent}. Using the inferred latent code, we query for the closest Fetch gripper grasp in the encoding space of training data grasps and execute it in the real world. We show a qualitative result of such a grasp transfer from a human demonstration in Fig.~\ref{fig:grasp-transfer}. More examples can be found in the supplementary materials. 

\begin{figure*}
	\centering
	\includegraphics[width=0.95\textwidth]{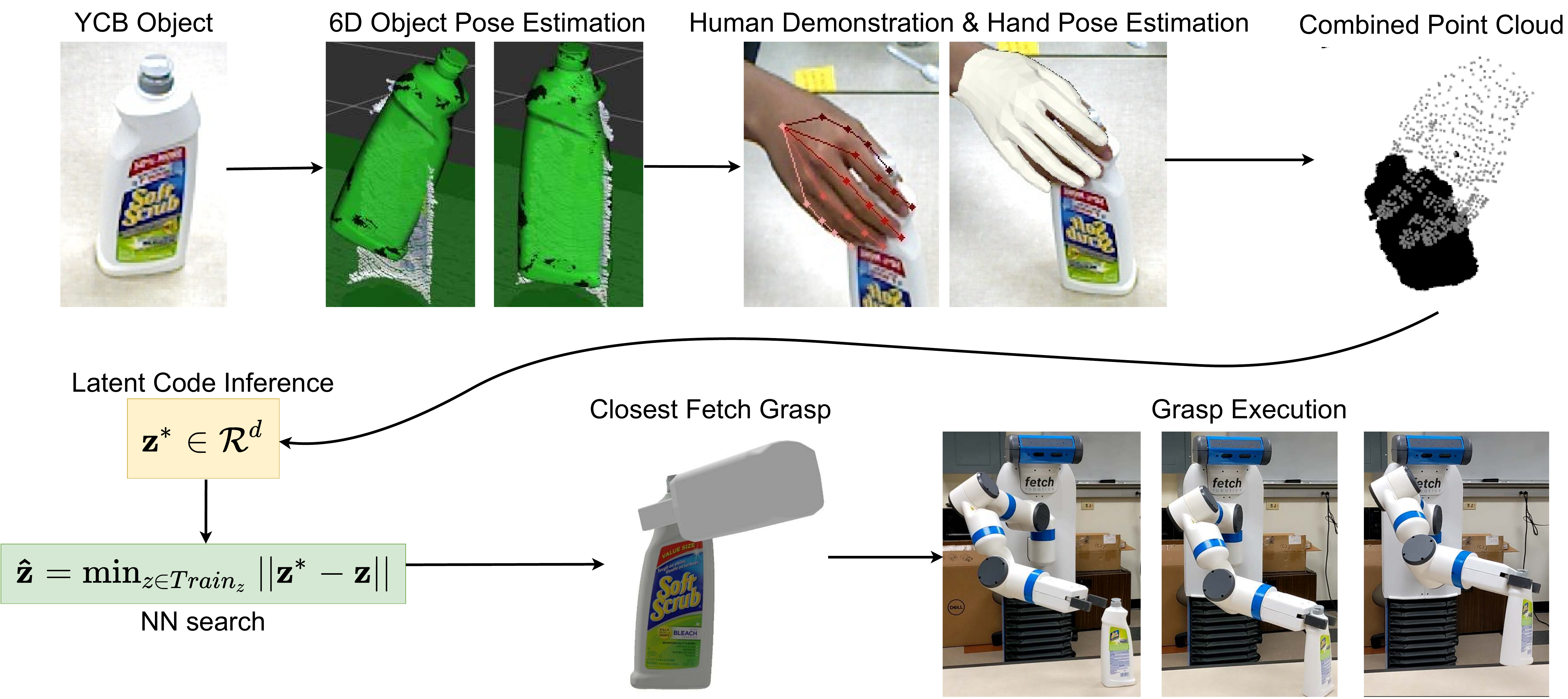}
	\caption{Our pipeline for grasp transfer from human demonstrations to a Fetch mobile manipulator.}
	\label{fig:grasp-transfer}
	\vspace{6mm}
\end{figure*}

\section{Conclusion and Future Work}
\label{sec:discussion}

We introduce NeuralGrasps, a framework for learning implicit representations for grasps of multiple robotic hands. In our framework, grasps across different robotic hands are embedded into a shared latent space, where the distance metric in the latent space is learned to preserve grasp similarity. Therefore, NeuralGrasps enables grasp transfer between grippers in the latent space. In addition, a latent code can be decoded to the signed distance functions of an object and a gripper in a grasping pose. This property enables us to perform 6D object pose estimation with grasping contact optimization using partial observations in the real world. With the estimated object pose, robot grasping can be achieved using the encoded grasps in NeuralGrasps.

In the future, we plan to address several limitations in NeuralGrasps. First, we train a single network for each object in our method. We will study learning representations of multiple objects and multiple grasps with a unified network. Second, NeuralGrasps cannot synthesize novel grasps in the latent space well. We will extend it to grasp synthesis in the future. Lastly, we will train NeuralGrasps with a large number of objects and make it be able to deal with unseen objects during inference.



\clearpage
\acknowledgments{This work was supported in part by the DARPA Perceptually-enabled Task Guidance (PTG) Program under contract number HR00112220005. B. Prabhakaran's work was supported by (while serving at) the National Science Foundation.}

\bibliography{main}  

\appendix

\vspace{4mm}
{\Large \textbf{Appendix}}

\section{Details about the Multi-Hand Grasping Dataset}
\label{sec:supp-dataset}
We use the Graspit!~\cite{miller2004graspit} simulation software to generate a set of
initial grasps for each gripper-object pair in a similar fashion as the ObMan
dataset~\cite{hasson2019learning}. Graspit assumes setting contact points on the gripper links which we manually set for 
grippers not present by default in Graspit (Franka Panda and Allegro grippers). We also ensure
to not select a proposed grasp if there is no contact with the object.
On the initial set of grasps generated by Graspit, we use farthest point 
sampling to obtain a final, fixed set of diverse grasps. In our experiments, 
the multi-hand grasping dataset has 60 grasps for each gripper-object pair i.e across the five grippers and the seven objects. The relevant grasp information (griper pose and configuration for the joints) generated by Graspit were stored as a json file which was later utilized in the PyBullet
rendering.
Once the grasps were generated, the object and gripper models were loaded in PyBullet 
environment with the pose and the joint configuration corresponding to the grasp applied on the 
gripper. We used 128 virtual Pybullet camera viewpoints around the object to ensure complete scene
information and render a dense point cloud for each grasping scene. 

Upon generation of the
point clouds, the SDF values for the gripper and the object were generated using the scheme 
outlined in~\cite{park2019deepsdf} with the caveat that we computed the SDFs over surface 
point clouds instead of 3D meshes as we had access to the surface points for the gripper and object
via the dense point cloud rendering.
For each object, we ensured that every grasping scene is scaled
to a unit sphere using a constant scaling factor across all grippers before the computing the
SDF training data to ensure that the relative differences in gripper sizes stay the same.
Given a desired number of points for the SDF generation (we used 40,000 points), there
was an equal division between points having (at least one of) \textit{positive} and 
\textit{negative} SDF values to either the gripper or the object. As seen in other related 
implicit representation works, a majority (95\%) of the sampled points 
were close to the gripper and object surface to encourage learning of SDF near their 
surfaces. The SDF values were also utilized in the contact map generation to find the object
contact points that were within 5cm of a gripper surface point. 

\section{Network Architecture and  Training Details} \label{sec:supp-model-training}
\begin{figure*}
	\centering
	\includegraphics[width=\textwidth]{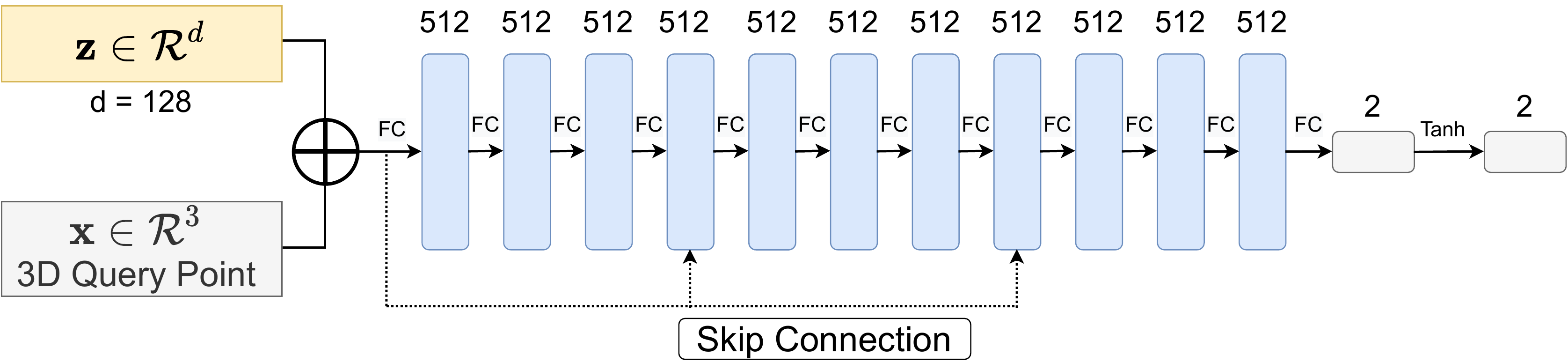}
	\caption{Network architecture used in our experiments. The 12 MLP blocks are shown along with the final layer tanh activation on the 2 output SDF values.}
	\label{fig:supp-network-layers}
\end{figure*}

The auto-decoder architecture shown in Fig.~\ref{fig:supp-network-layers} is used in all of our experiments. The decoder 
component consists of 12 fully connected layers with each having 512 dimensions, and finally 
the two SDF values (for gripper and object) as the output. We use Batch 
Normalization ~\cite{ioffe2015batchnorm} and Dropout 
~\cite{srivastava2014dropout} after each layer with a dropout probability 
of 0.2.
The query point 
and latent code for a specific grasping scene are concatenated and passed as 
input to the auto-decoder. The latent code size (\(d\)) for all experiments 
was 128 and the latent code distribution had a zero-mean Gaussian prior 
\(\mathcal{N}(0, \sigma^2 \mathcal{I}_d)\) with \(\sigma = 0.01\). 
We used the Adam~\cite{kingma2014adam} optimizer to the train the network 
parameters and the latent codes for a total of 2,000 epochs with a step 
learning rate schedule for the optimizer.
From the 60 grasps for every object-gripper pair in the dataset, we choose 50 of them across
each object to construct the training set for the neural network model training. 
For the loss function on the SDF 
values, we have equal weight between the individual components on the gripper, object SDFs and the
triplet loss. Each component was given a weight of 1.0. The SDF clamping distance 
\(\delta\) is set to be 0.05 and for each grasping scene, we sample 20,000 points from the corresponding SDF samples in the dataset.

\section{Experiment Details}
\subsection{Shape Reconstruction and Grasp Retrieval}
For the experiments on unseen test grasps mentioned in the paper, we pick
a set of held-out grasps not used during the model training. For an object,
we select 5 grasping scenes from every gripper as the test set.
The latent code for each
test grasp is inferred using the SDF samples prepared during dataset
construction and doing the MAP optimization for 800 iterations. Once the
latent code is inferred, we try to predict the SDF values (query it via the
model) on a randomly sampled set of query points inside the unit sphere. For all our experiments, we sample 100,000 points inside the unit sphere. 
Using the predicted SDF values on the query points, we obtain the estimated
point clouds for the object and the gripper as the query points having the
SDFs less than zero for object and gripper respectively. Chamfer distance
is then computed between the inferred and ground truth point clouds for the
object and gripper in the specific test grasping scene. 

In the grasp retrieval experiments, for an input query grasp, the \(L_1\) 
distance matrices for 1) the inferred latent code of the query grasp and training grasp
latent codes and 2) the ground truth contact map of the query grasp and training set contact
maps, each has the distances normalized to be between 0 and 1 so that the 
similarity score (computed as 1-distance) also remains between 0 and 1. We opted to use the simple 1-distance measure since any monotonically decreasing function of the distance works as a notion of similarity. Once we have the 
distances between the query and training set grasps, we simply sort them to
obtain the \textit{closest} and \textit{farthest} examples.

\begin{figure*}
	\centering
	\includegraphics[width=\textwidth]{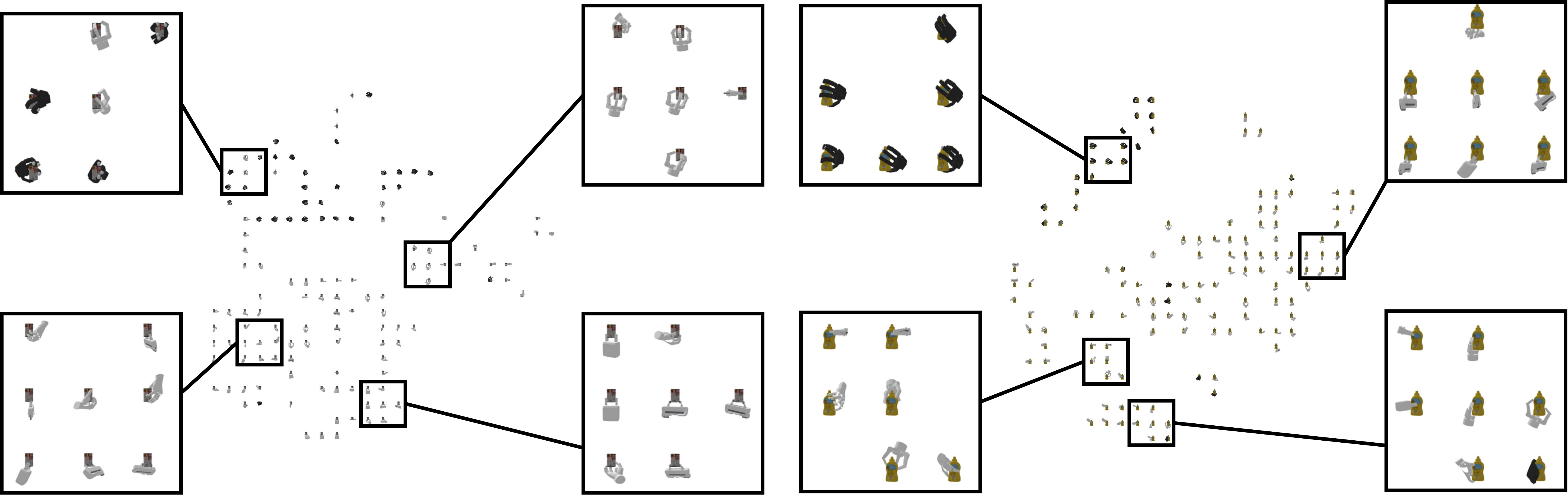}
	\caption{t-SNE visualization of the grasp embeddings on tomato soup can (left) and mustard bottle (right).}
	\label{fig:supp-tsne-005-006}
\end{figure*}
\begin{figure*}
	\centering
	\includegraphics[width=\textwidth]{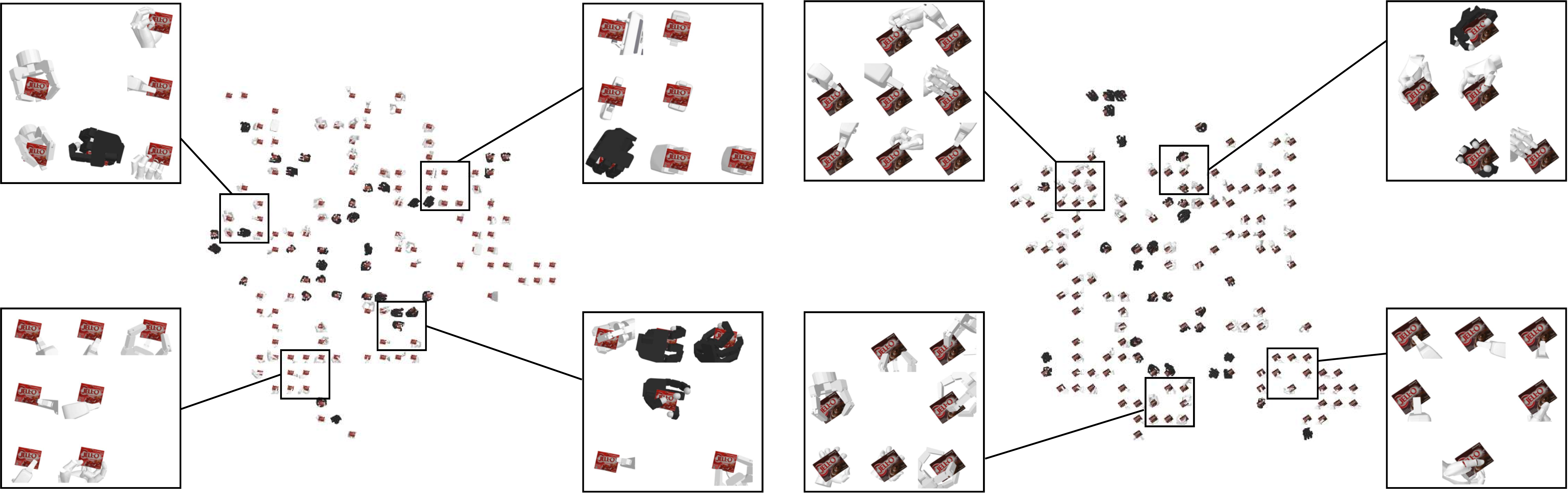}
	\caption{t-SNE visualization of the grasp embeddings on pudding box (left)and gelatin box (right).}
	\label{fig:supp-tsne-008-009}
\end{figure*}
\begin{figure*}
	\centering
	\includegraphics[width=0.5\textwidth]{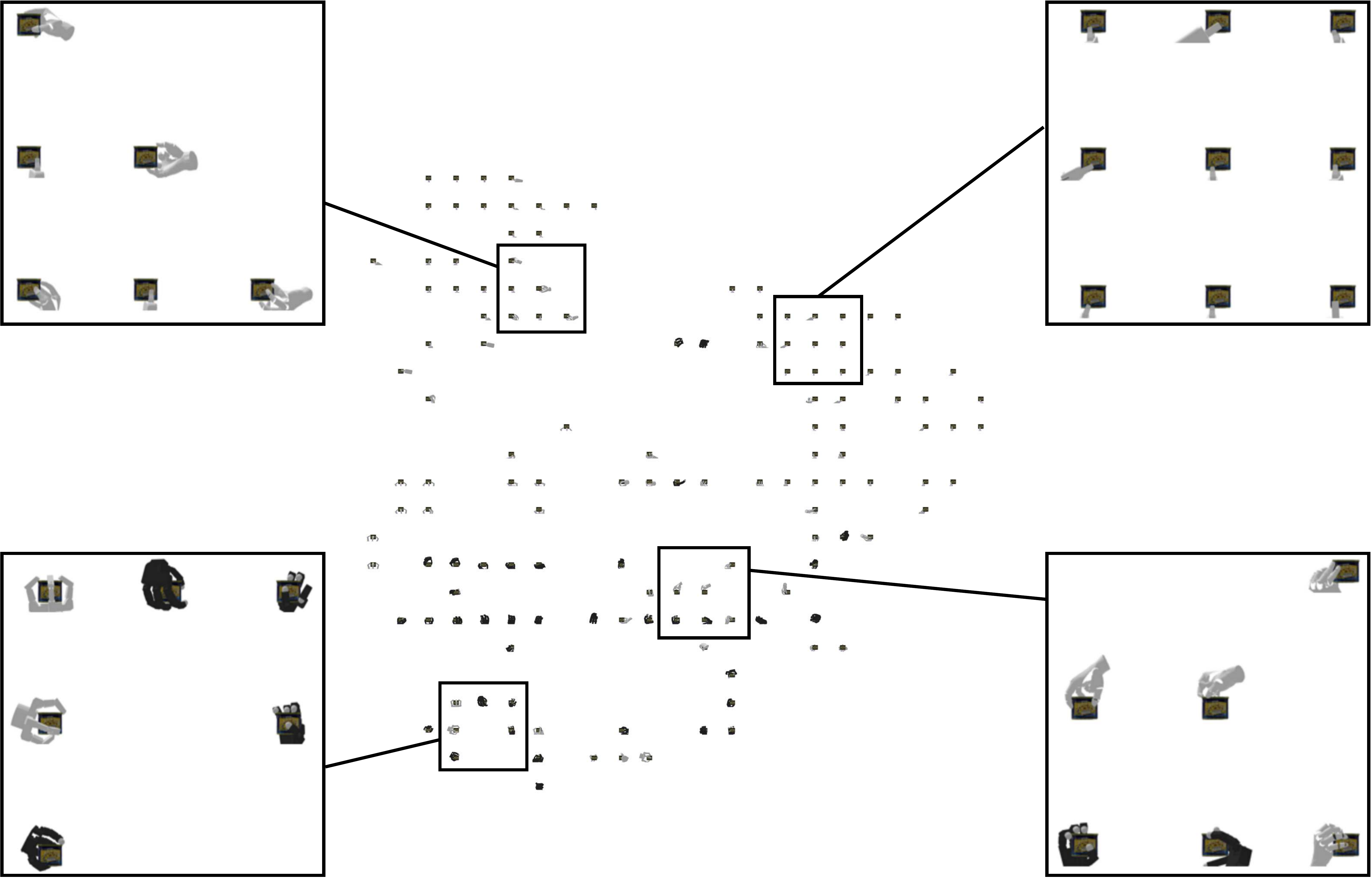}
	\caption{t-SNE visualization of the grasp embeddings on potted meat can.}
	\label{fig:supp-tsne-010}
\end{figure*}

We present the Chamfer distances and grasp retrieval similarity scores on the grasps in the 
training set in Table~\ref{tab:supp-exp-cdist-retrieval-train}. The training set experiments 
were used as a sanity check on the model for representing grasps it had already seen. 
Although reconstruction is not the primary goal, we see low values for 
Chamfer distance and a high degree of match between the ground truth contact map-based and latent code-based grasp retrieval results. The Barnes-Hut t-SNE visualizations~\cite{van2014accelerating} of the learned embeddings of the different grasps from five robotic hands across the different objects are shown in Figs.~\ref{fig:supp-tsne-005-006},~\ref{fig:supp-tsne-008-009},~\ref{fig:supp-tsne-010}.

\begin{table}[!ht]
\caption{Chamfer distance (x1e-3) for shape reconstruction and contact map similarity for grasp retrieval on training set.}
\label{tab:supp-exp-cdist-retrieval-train}
\centering
\resizebox{0.99\linewidth}{!}{
\begin{tabular}{|c|cccccc|cccc|}
\hline
Ycb Model &
  \multicolumn{6}{c|}{Shape Reconstruction} &
  \multicolumn{4}{c|}{Grasp Retrieval} \\ \hline
 &
  \multicolumn{1}{c|}{Object} &
  \multicolumn{1}{c|}{Fetch} &
  \multicolumn{1}{c|}{Barrett} &
  \multicolumn{1}{c|}{Human} &
  \multicolumn{1}{c|}{Allegro} &
  \multicolumn{1}{c|}{Panda} &
  \multicolumn{1}{c|}{Near Z} &
  \multicolumn{1}{c|}{Near GT} &
  \multicolumn{1}{c|}{Far Z} &
  \multicolumn{1}{c|}{Far GT} \\ \cline{2-11} 
\textit{Cracker Box} &
  \multicolumn{1}{c|}{1.28} &
  \multicolumn{1}{c|}{4.06} &
  \multicolumn{1}{c|}{5.89} &
  \multicolumn{1}{c|}{6.36} &
  \multicolumn{1}{c|}{4.11} &
  7.84 &
  \multicolumn{1}{c|}{0.87} &
  \multicolumn{1}{c|}{0.88} &
  \multicolumn{1}{c|}{0.26} &
  0.24 \\
\textit{Soup Can} &
  \multicolumn{1}{c|}{1.29} &
  \multicolumn{1}{c|}{3.77} &
  \multicolumn{1}{c|}{4.84} &
  \multicolumn{1}{c|}{2.91} &
  \multicolumn{1}{c|}{3.30} &
  3.12 &
  \multicolumn{1}{c|}{0.78} &
  \multicolumn{1}{c|}{0.79} &
  \multicolumn{1}{c|}{0.12} &
  0.10 \\
\textit{Mustard Bottle} &
  \multicolumn{1}{c|}{0.99} &
  \multicolumn{1}{c|}{1.91} &
  \multicolumn{1}{c|}{3.22} &
  \multicolumn{1}{c|}{2.14} &
  \multicolumn{1}{c|}{1.98} &
  2.48 &
  \multicolumn{1}{c|}{0.83} &
  \multicolumn{1}{c|}{0.84} &
  \multicolumn{1}{c|}{0.19} &
  0.16 \\
\textit{Pudding Box} &
  \multicolumn{1}{c|}{1.22} &
  \multicolumn{1}{c|}{5.14} &
  \multicolumn{1}{c|}{8.16} &
  \multicolumn{1}{c|}{6.55} &
  \multicolumn{1}{c|}{5.25} &
  6.11 &
  \multicolumn{1}{c|}{0.83} &
  \multicolumn{1}{c|}{0.84} &
  \multicolumn{1}{c|}{0.24} &
  0.23 \\
\textit{Gelatin Box} &
  \multicolumn{1}{c|}{1.12} &
  \multicolumn{1}{c|}{5.53} &
  \multicolumn{1}{c|}{7.73} &
  \multicolumn{1}{c|}{4.66} &
  \multicolumn{1}{c|}{4.99} &
  4.73 &
  \multicolumn{1}{c|}{0.80} &
  \multicolumn{1}{c|}{0.81} &
  \multicolumn{1}{c|}{0.23} &
  0.21 \\
\textit{Potted Meat Can} &
  \multicolumn{1}{c|}{1.96} &
  \multicolumn{1}{c|}{3.90} &
  \multicolumn{1}{c|}{5.23} &
  \multicolumn{1}{c|}{3.78} &
  \multicolumn{1}{c|}{3.77} &
  3.05 &
  \multicolumn{1}{c|}{0.79} &
  \multicolumn{1}{c|}{0.80} &
  \multicolumn{1}{c|}{0.14} &
  0.10 \\
\textit{Bleach Cleanser} &
  \multicolumn{1}{c|}{7.09} &
  \multicolumn{1}{c|}{4.87} &
  \multicolumn{1}{c|}{6.12} &
  \multicolumn{1}{c|}{7.52} &
  \multicolumn{1}{c|}{5.50} &
  8.85 &
  \multicolumn{1}{c|}{0.72} &
  \multicolumn{1}{c|}{0.73} &
  \multicolumn{1}{c|}{0.18} &
  0.17 \\ \hline
\end{tabular}}
\end{table}

\subsection{Robotic Grasping}

For the object pose estimation and grasping experiments, we assume an
uncluttered tabletop scene with a single object placed on the table. Since the
goal of the experiments is not to showcase detection or segmentation on object
point clouds, we adopt a simple heuristic of using the table height to segment
out the real world point cloud of the target object. We only use a single-view (partial)
point cloud of the object in the real-world experiments. 

\textbf{Baseline Model for 6D Object Pose Estimation.} The pose optimization
assumes that the given points lie on the object surface and hence the predicted
SDF values for them should be close to zero (first term of Equation~\ref{eq:6d-pose-contact}). We solve the following optimization problem to estimate the 3D rotation $R$ and the 3D translation $\mathbf{t}$ of the object pose:
\begin{equation} \label{eq:6d-pose-baseline}
    R^*, \mathbf{t}^* = \arg \min_{ R, \mathbf{t}} \Big [ \sum_{j=1}^M  | f_{\text{SDF}}^o(R \mathbf{x}_c^j + \mathbf{t}, \mathbf{z}; \theta_o) | \Big ],
\end{equation}
where $\{ \mathbf{x}_c^j \}_{j=1}^M$ denotes the points in the camera frame. $\mathbf{z}$ is the latent code of an arbitrary grasp in the training set since we do not use the SDF of the gripper in this optimization. We name this model the baseline model for object pose estimation since it does not utilize the grasp contact. To start the optimization with a good initial pose, we initialize the 3D translation with the center of the point cloud. For 3D rotation, we discretize the SO(3) space by yaw, pitch and roll angles with 45-degree intervals, then evaluate and randomly select a 3D rotation within the top-5 minimum objective values according to Eq.~\eqref{eq:6d-pose-baseline} as the initial rotation. After the optimization, we accept the estimated pose if the objective value is smaller than a pre-defined threshold. Otherwise, we re-run the optimization until a good pose is obtained. These steps are useful to avoid local minimums in SDF-based optimization for object pose estimation.

\textbf{6D Object Pose Estimation with Contact.}  Let's denote the estimated object pose using the baseline model as $(R_0, \mathbf{t}_0)$. Using this initial pose estimation, we find a Fetch gripper grasp 
from the training set closest to the current gripper position for optimization. Because all the grasps are defined in the coordinate frame of the object. We need to use the object pose to transform the grasps to the camera frame, and use the pose between camera and robot base to transform them into the robot base frame. After that, we can compare them with the current gripper pose in the robot base frame. Assume the select grasp has a latent code $\mathbf{z}^*$ and contact map $\phi$. We can first find a set of contact points on the object surface $\{ \mathbf{x}_o^k \}_{k=1}^L$ using the contact map $\phi$. Note that these points are defined in the object coordinate frame. Using the initial pose estimation $(R_0, \mathbf{t}_0)$, we can transform these points to the camera frame and then find the nearest neighbors of the transformed points from the object point cloud in the camera frame. Let's denote these nearest neighbor points in the camera frame as $\{ \mathbf{x}_g^k \}_{k=1}^L$. We consider these points to be the contact points in the camera frame for the grasp encoded by $\mathbf{z}^*$. Therefore, we can perform the following optimization to refine the initial pose:
\begin{equation} \label{eq:6d-pose-contact}
    R^*, \mathbf{t}^* = \arg \min_{ R, \mathbf{t}} \Big [ \sum_{j=1}^M  | f_{\text{SDF}}^o(R \mathbf{x}_c^j + \mathbf{t}, \mathbf{z}^*; \theta_o) | + \sum_{k=1}^L | f_{\text{SDF}}^g(R \mathbf{x}_g^k + \mathbf{t}, \mathbf{z}^*; \theta_o) |	 \Big ],
\end{equation}
This is our algorithm for 6D object pose estimation with contact using our implicit representations.


\textbf{Grasp Execution.} In order to execute the chosen grasp with optimized object pose, we first compute a
standoff position and then sample 10 way points from the the standoff to the 
final grasp position for smooth execution. In the videos, it can be seen that
sometimes the motion from standoff to final pose is very fast resulting in
some momentum transferred to the object. We add the table on which
the object is placed as an obstacle and then utilize the 
Moveit~\cite{chitta2012moveit} package to do the motion planning from an initial
stowed position for the Fetch gripper. Lastly, on arriving in the final position, the
Fetch gripper attempts to grasp the object by closing its fingers and then
lifting the object from the table.
The pose estimation method is not fail-proof as seen from the results
of Table 2 in the paper and attached video containing the grasping clips. A few
failures are a consequence of the pose estimation getting stuck in a local minima. In
some cases, the pose estimation for the \textit{``with contact''} method actually
performed worse due to an incorrect closest point matching, and likely resulting from 
a bad initial pose estimate.

\subsection{Human Demonstrations and Grasp Transfer}
We utilize existing methods for hand pose estimation in our human-to-robot
grasp transfer experiments. The same RGB-D robot head camera is used to observe
the human grasp demonstrations and infer the hand pose and 3D mesh for it on a 
per frame basis. Following a demo of the desired grasp by a human hand, we estimate the human 
hand grasps with the following pipeline. Given a RGB-D video of the human grasp demo, 
the Region of Interest (RoI) around the hand is detected using the Fully Convolutional One-Stage Object Detector (FCOS~\cite{tian2019fcos}) trained on 100 Days of Hands~\cite{shan2020understanding}
dataset. Next, the RoI of the hand is cropped from the depth image, and the 3D hand joints are estimated using the Anchor to Joint (A2J ~\cite{xiong2019a2j}) 
model trained on the DexYCB~\cite{chao2021dexycb} dataset. Lastly, we use the 
Pose2Mesh~\cite{choi2020pose2mesh} model trained on FreiHAND~\cite{zimmermann2019freihand} dataset to 
recover a 3D mesh of the hand from the predicted 3D hand joints. The 3D mesh consequently also gives the 
hand surface point cloud required for the latent code inference (here the hand is considered as a 
\textit{gripper}).
\begin{figure*}[!ht]
	\centering
	\includegraphics[width=\textwidth]{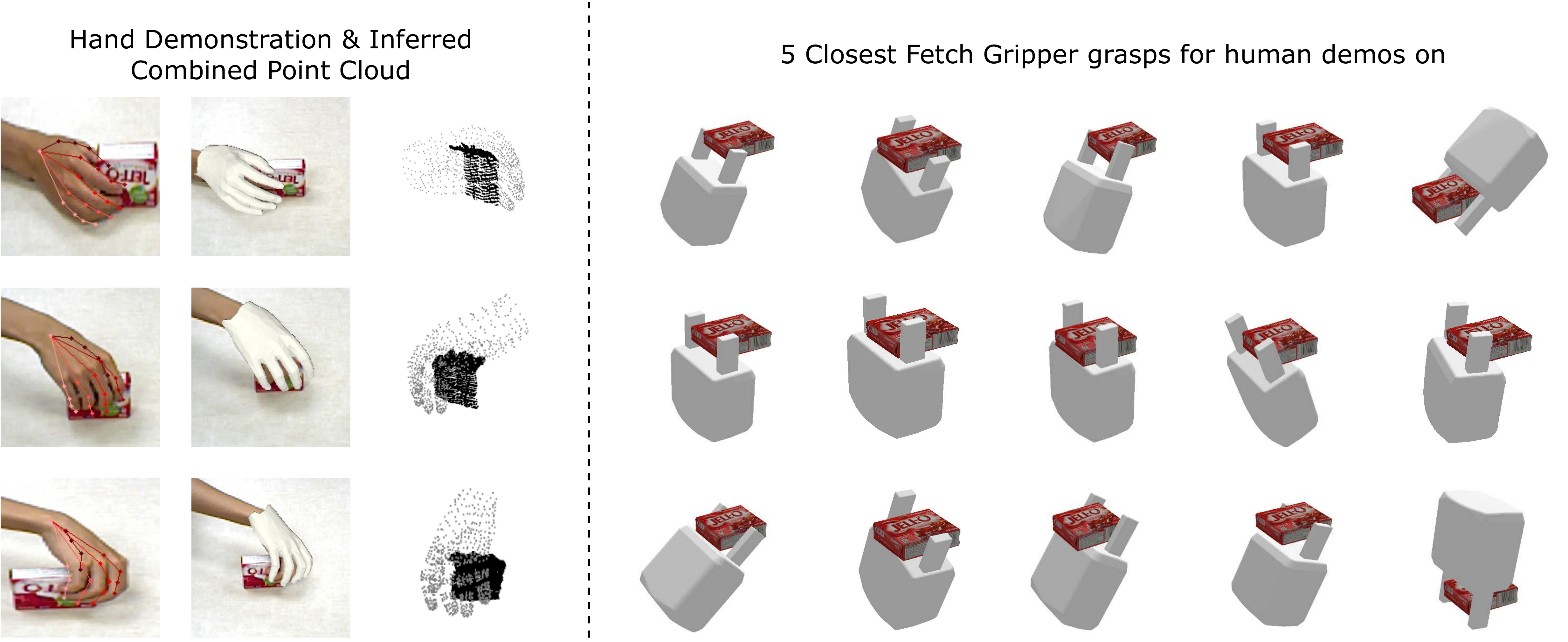}
	\caption{Closest Fetch gripper grasps for the human demonstrations on the gelatin box (009). The grasps in the dataset generated from Graspit are not similar to a pincer like grasp.}
	\label{fig:supp-closest_Fetch_grasps_grippers-009}
\end{figure*}
The supplementary video also shows how the grasp transfer process looks like with
demos for each object. Once we have the closest Fetch gripper grasps, the Moveit~\cite{chitta2012moveit}
motion planning package is used as before to find a motion plan to the closest grasp. We note
that the closest retrieved Fetch grasp might not have a feasible motion plan associated with it and hence
we iterated along the ordered list of closest grasps and execute the one for which a successful motion
plan is found. This set of executed grasps are shown in the video and hence for a few cases, the
execution slightly differs from the demonstration. 

\begin{figure*}
	\centering
	\includegraphics[width=\textwidth]{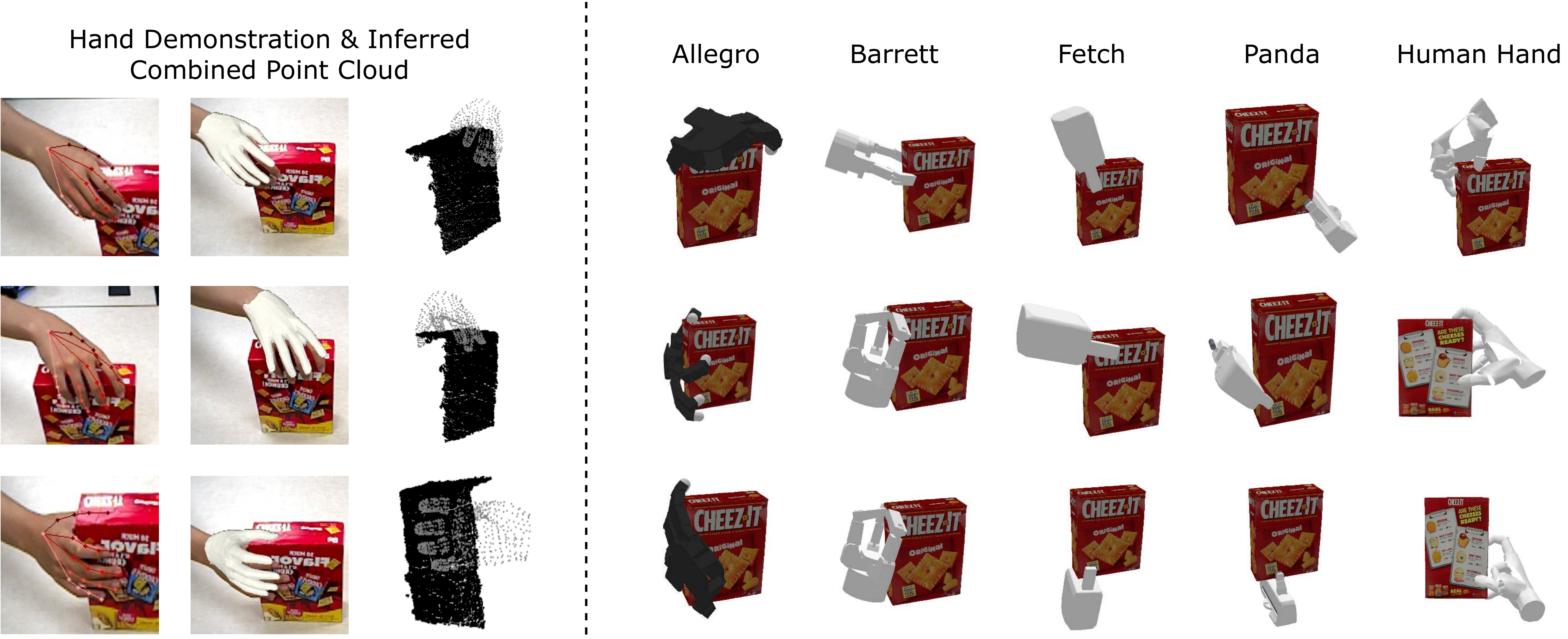}
	\includegraphics[width=\textwidth]{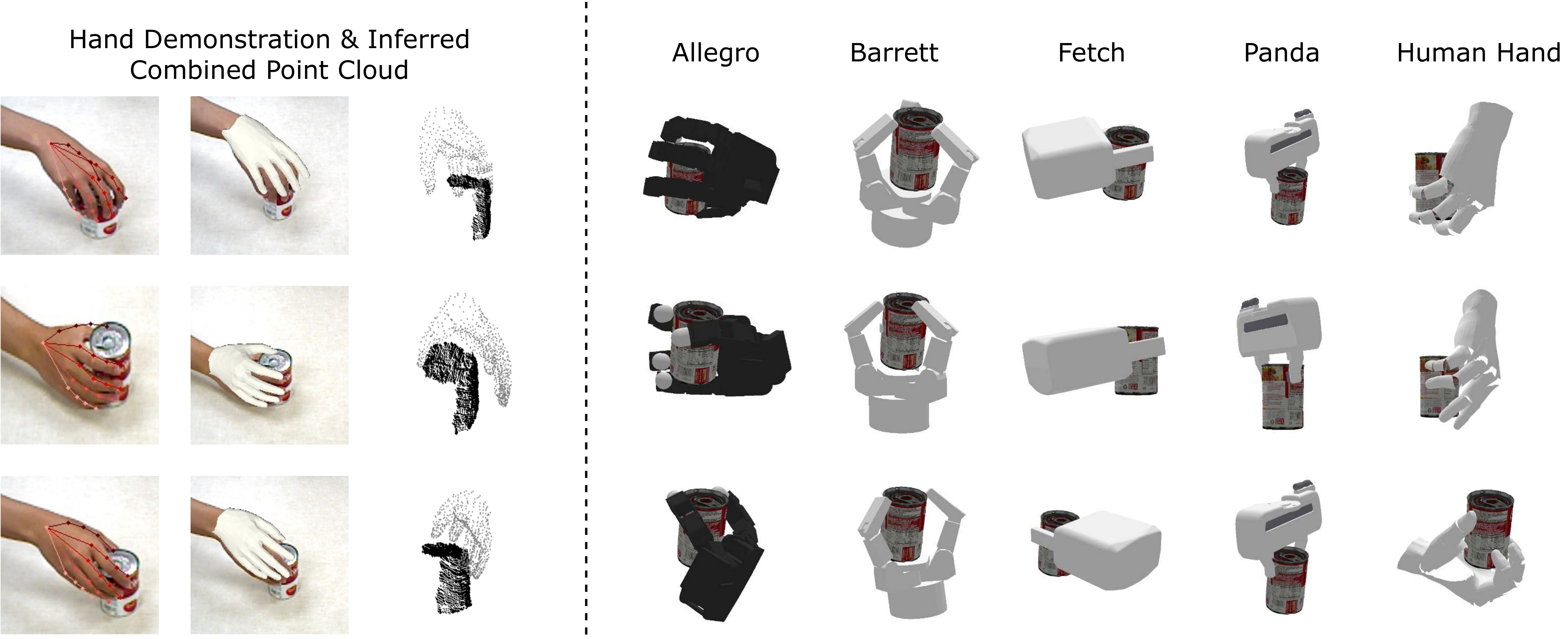}
	\caption{Closest grasps for all grippers for the human demonstrations on the cracker box (top) and tomato soup can (bottom)}
	\label{fig:supp-closest_grasp_all_grippers-003-005}
\end{figure*}

\begin{figure*}
	\centering
	\includegraphics[width=\textwidth]{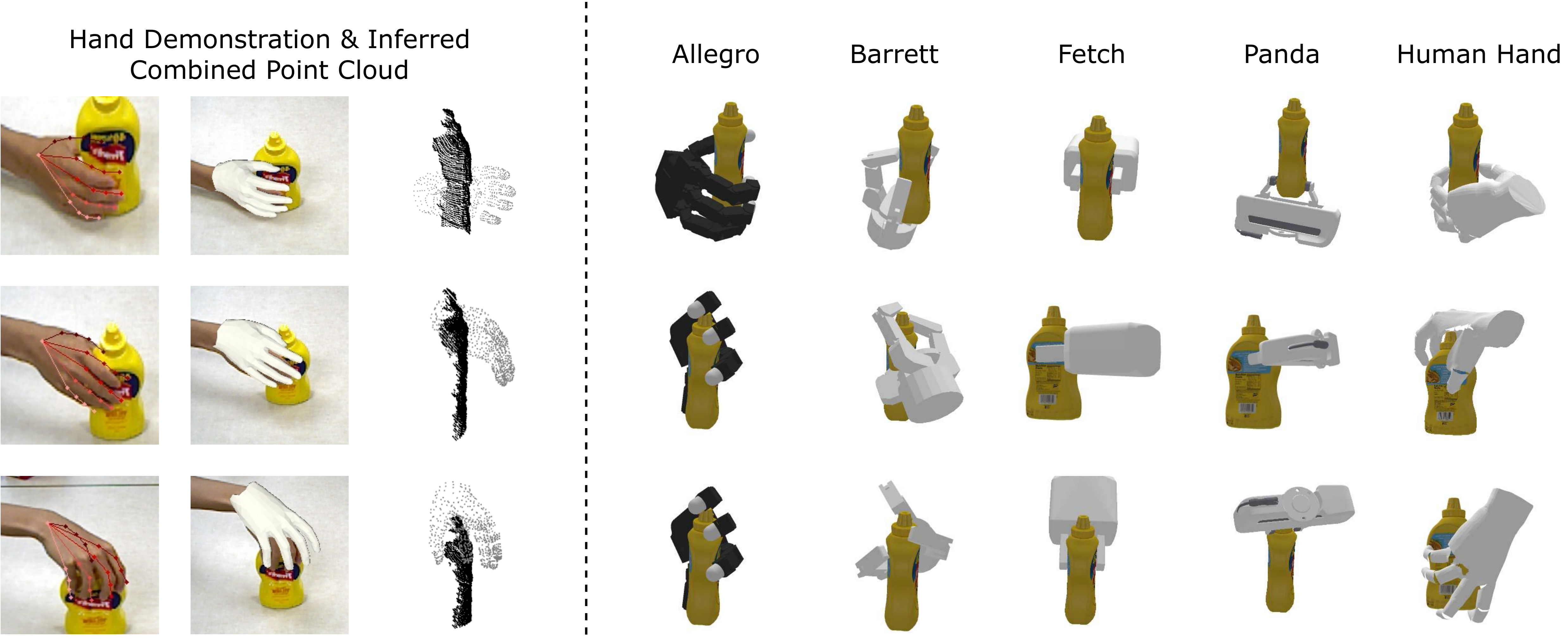}
	\includegraphics[width=\textwidth]{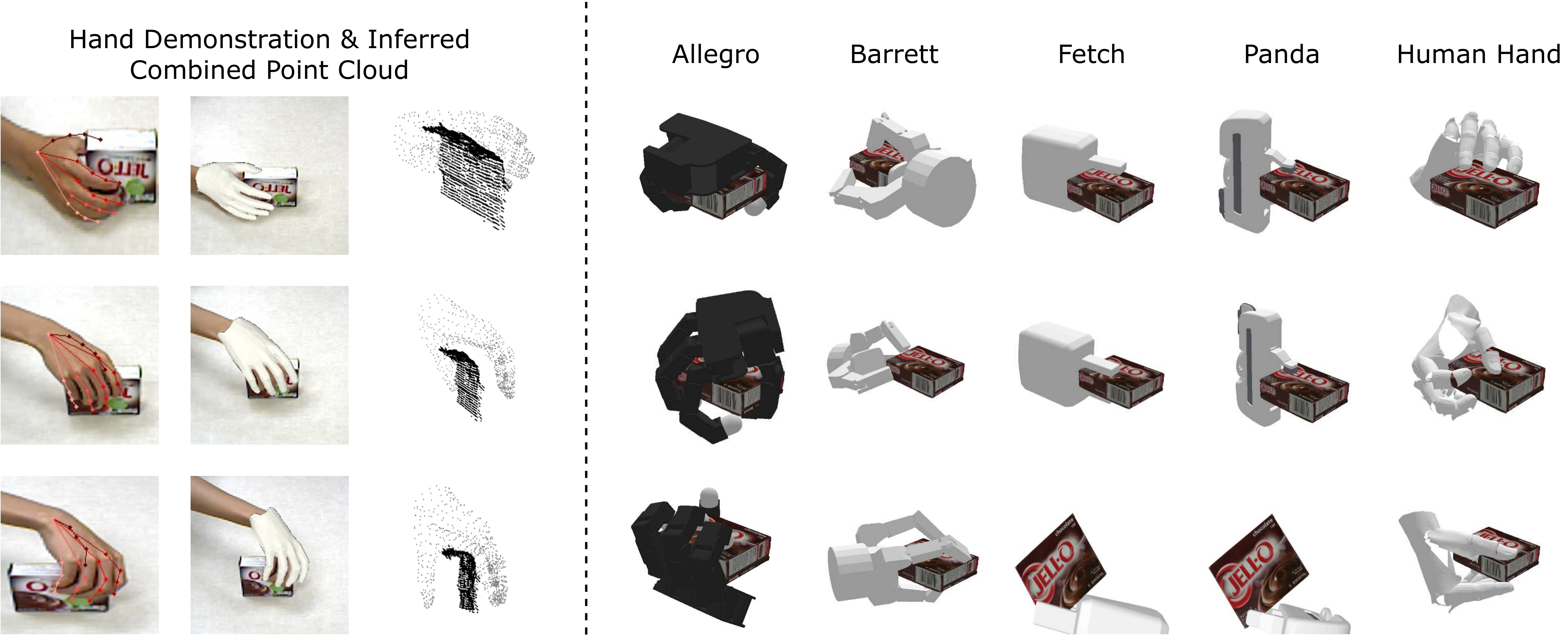}
	\includegraphics[width=\textwidth]{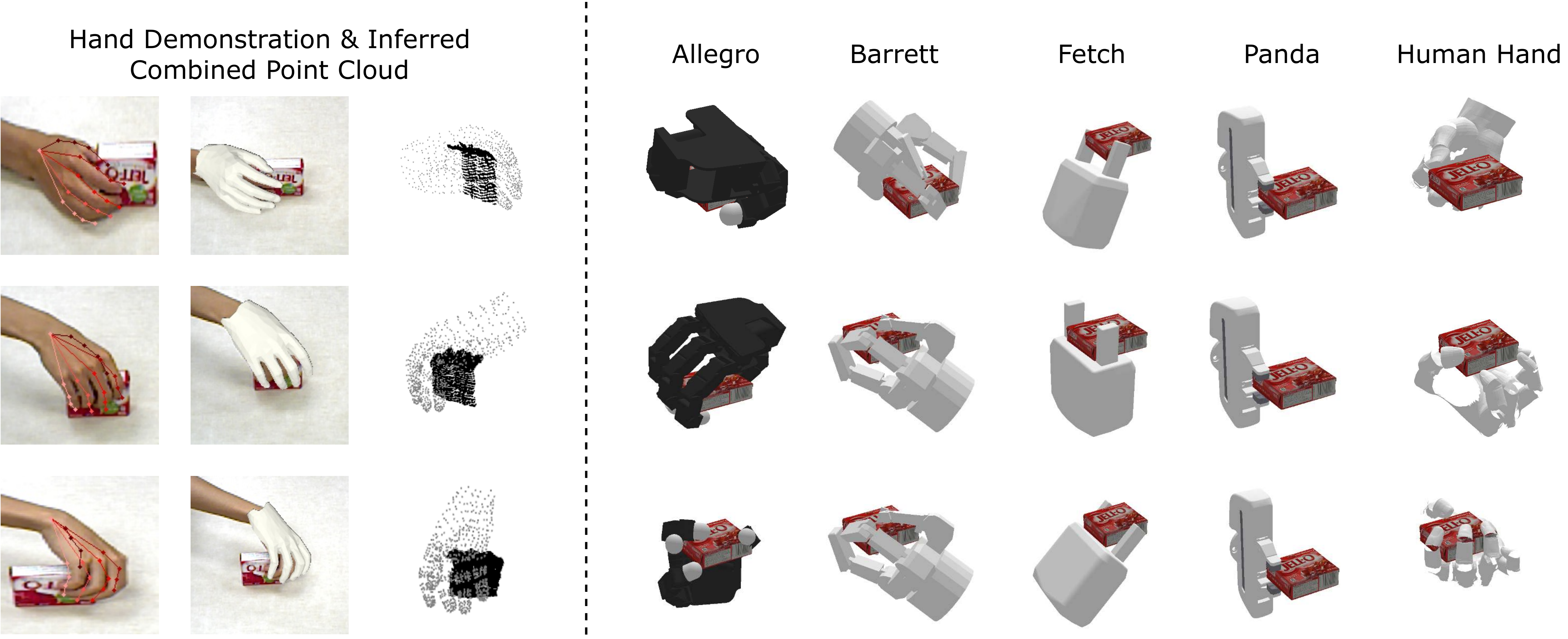}
	\caption{Closest grasps for all grippers for the human demonstrations on the mustard bottle (top), 
	 pudding box (middle), and gelatin box (bottom).}
	\label{fig:supp-closest_grasp_all_grippers-006-008-009}
\end{figure*}

\begin{figure*}
	\centering
	\includegraphics[width=\textwidth]{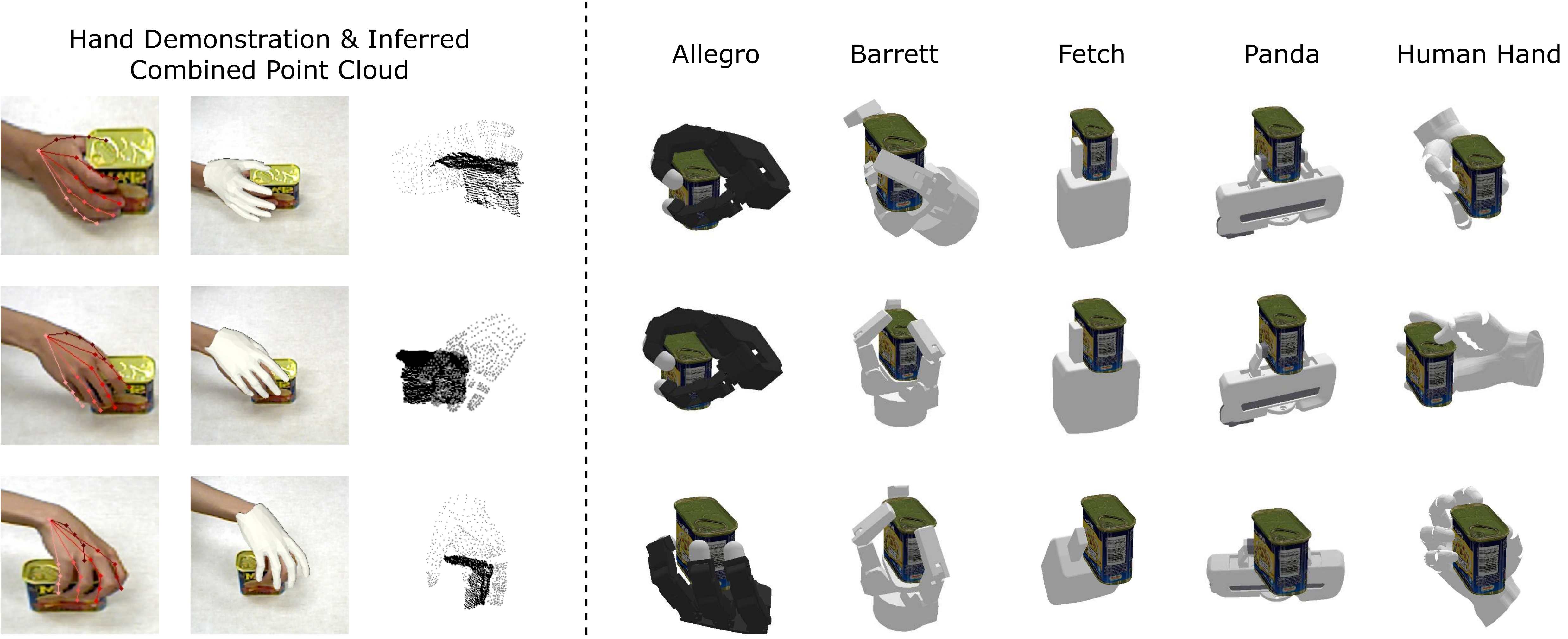}
	\includegraphics[width=\textwidth]{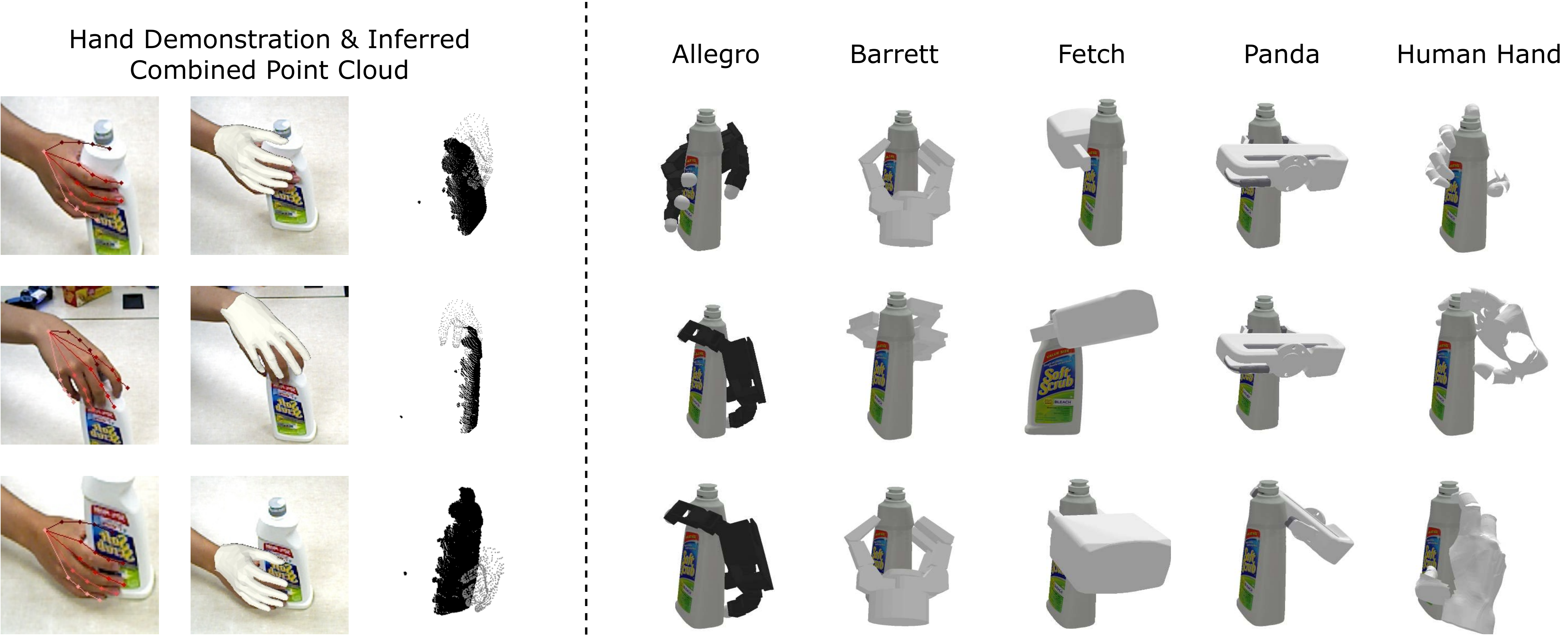}
 	\vspace{7mm}
	\caption{Closest grasps for all grippers for the human demonstrations on the potted meat can (top) and bleach cleanser (bottom).}
	\label{fig:supp-closest_grasp_all_grippers-010-021}
\end{figure*}

The grasp transfer for the gelatin box (009) demo was not successful due to the motion 
planning finding no feasible plan to the inferred closest Fetch gripper's grasp. This is 
likely due to the fact that the inferred closest Fetch grippers grasps have the gripper to grasp the 
box using the top and bottom rather than the corner as see in Fig.~\ref{fig:supp-closest_Fetch_grasps_grippers-009}.
Additional qualitative results on closest inferred grasps across the entire gripper set are shown in 
Fig.~\ref{fig:supp-closest_grasp_all_grippers-003-005}, ~\ref{fig:supp-closest_grasp_all_grippers-006-008-009}, and~\ref{fig:supp-closest_grasp_all_grippers-010-021}.
It is possible that the transfer process is sometimes not feasible for such
scenarios where the mapping to the Fetch gripper is not perfect. It can be a consequence of the dataset 
not containing a specific type of Fetch gripper grasp and the existing grasps being very different from
the human demo even though the contact regions may be close on a small object relative to the grippers like the gelatin box.
Another avenue to consider in the grasp transfer process is utilizing reliable object pose estimates of real world scans. For
example in Fig.~\ref{fig:supp-closest_grasp_all_grippers-003-005}, for the first row of cracker box demonstration, the 
closest Panda gripper grasp is diagonally opposite to others. This result is due to the fact that the
latent code inference does not take texture information into account and hence we have a flipped result. 

\end{document}